\documentclass[12pt,twoside]{article} 
 
 \usepackage[pdftex]{graphicx}
\usepackage{algorithm}
\usepackage{algorithmic}
\date{} 
\begin{document} 
\centerline{\bf Applied Mathematical Sciences, Vol. 6, 2012, no. 102, 5085 - 5095} 
\centerline{} 
\centerline{} 
\centerline {\Large{\bf Dimensionality Reduction and Classification Feature}} 
\centerline{} 
\centerline{\Large{\bf Using Mutual Information Applied to Hyperspectral  }} 
\centerline{}
\centerline{\Large{\bf Images: A Filter Strategy Based Algorithm}} 
\centerline{} 
\centerline{\bf {ELkebir Sarhrouni*, Ahmed Hammouch** and Driss Aboutajdine*}} 
\centerline{} 
\centerline{*LRIT, Faculty of Sciences, Mohamed V - Agdal University, Morocco} 
\centerline{**LRGE, ENSET, Mohamed V - Souissi University, Morocco} 

\centerline{sarhrouni436@yahoo.fr, hammouch\_a@yahoo.com, aboutaj@fsr.ac.ma} 

\newtheorem{Theorem}{\quad Theorem}[section] 
\newtheorem{Definition}[Theorem]{\quad Definition} 
\newtheorem{Corollary}[Theorem]{\quad Corollary} 
\newtheorem{Lemma}[Theorem]{\quad Lemma} 
\newtheorem{Example}[Theorem]{\quad Example} 

\begin{abstract} 
Hyperspectral images (HIS) classification is a high technical remote sensing tool.
The goal is to reproduce a thematic map that will be compared with a reference
ground truth map (GT), constructed by expecting the region. The HIS contains
more than a hundred bidirectional measures, called bands (or simply images), of
the same region. They are taken at juxtaposed frequencies. Unfortunately, some
bands contain redundant information, others are affected by the noise, and the
high dimensionality of features made the accuracy of classification lower. The
problematic is how to find the good bands to classify the pixels of regions. Some
methods use Mutual Information (MI) and threshold, to select relevant bands,
without treatment of redundancy. Others control and eliminate redundancy by
selecting the band top ranking the MI, and if its neighbors have sensibly the same
MI with the GT, they will be considered redundant and so discarded. This is the
most inconvenient of this method, because this avoids the advantage of
hyperspectral images: some precious information can be discarded. In this paper
we’ll accept the useful redundancy. A band contains useful redundancy if it
contributes to produce an estimated reference map that has higher MI with the GT.
To control redundancy, we introduce a complementary threshold added to last
value of MI. This process is a Filter strategy; it gets a better performance of  classification accuracy and not expensive, but less preferment than Wrapper
strategy.
\end{abstract} 

{\bf Keywords:} Hyperspectral images, classification, feature selection, mutual
information, redundancy

\section{Introduction} 
In the feature classification domain, the choice of data affects widely the results.
For the Hyperspectral image, the bands don’t all contain the information; some
bands are irrelevant like those affected by various atmospheric effects, see
Figure.4, and decrease the classification accuracy. And there exist redundant
bands to complicate the learning system and product incorrect prediction [14].
Even the bands contain enough information about the scene they may can’t
predict the classes correctly if the dimension of space images, see Figure.3, is so
large that needs many cases to detect the relationship between the bands and the
scene (Hughes phenomenon) [10]. We can reduce the dimensionality of
hyperspectral images by selecting only the relevant bands (feature selection or
subset selection methodology), or extracting, from the original bands, new bands
containing the maximal information about the classes, using any functions, logical
or numerical (feature extraction methodology) [11][9]. Here we focus on the
feature selection using mutual information. Hyperspectral images have three
advantages regarding the multispectral images [6], see  Figure 1.
{\bf First:} the hyperspectral image contains more than a hundred images but the
multispectral contains three at ten images.\\
{\bf Second:} hyperspectral image has a spectral resolution (the central wavelength
divided by de width of spectral band) about a hundred, but that of multispectral is
about ten.\\
{\bf Third:} the bands of a hyperspectral image is regularly spaced, those of
multispectral image is large and irregularly spaced.\\
\textit{{\bf Assertion:} when we reduce hyperspectral images dimensionality, any method
used must save the precision and high discrimination of substances given by
hyperspectral image.}

\begin{figure}[!h]
\centering
\includegraphics[width=3.5in]{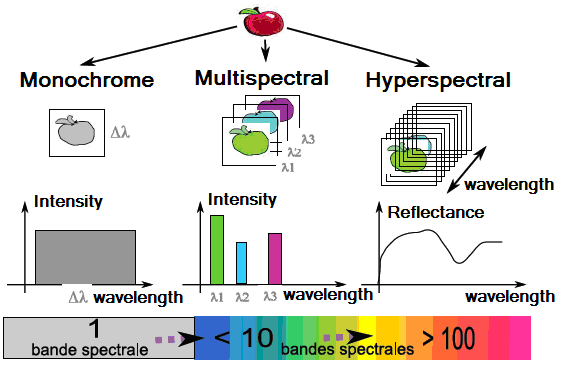}
\caption{Precision an dicrimination added by hyperspectral images}
\label{fig_sim}
\end{figure}

In this paper we use the Hyperspectral image AVIRIS 92AV3C (Airborne Visible
Infrared Imaging Spectrometer). [2]. It contains 220 images taken on the region
"Indiana Pine" at "north-western Indiana", USA [1]. The 220 called bands are
taken between 0.4μm and 2.5μm. Each band has 145 lines and 145 columns. The
ground truth map is also provided, but only 10366 pixels are labeled fro 1 to 16.
Each label indicates one from 16 classes. The zeros indicate pixels how are not
classified yet, Figure.2.\\
The hyperspectral image AVIRIS 92AV3C contains numbers between 955 and
9406. Each pixel of the ground truth map has a set of 220 numbers (measures)
along the hyperspectral image. This numbers (measures) represent the reflectance
of the pixel in each band. So the pixel is shown as vector off 220 components.
Figure .3.

\begin{figure}[!th]
\centering
\includegraphics[width=3.5in]{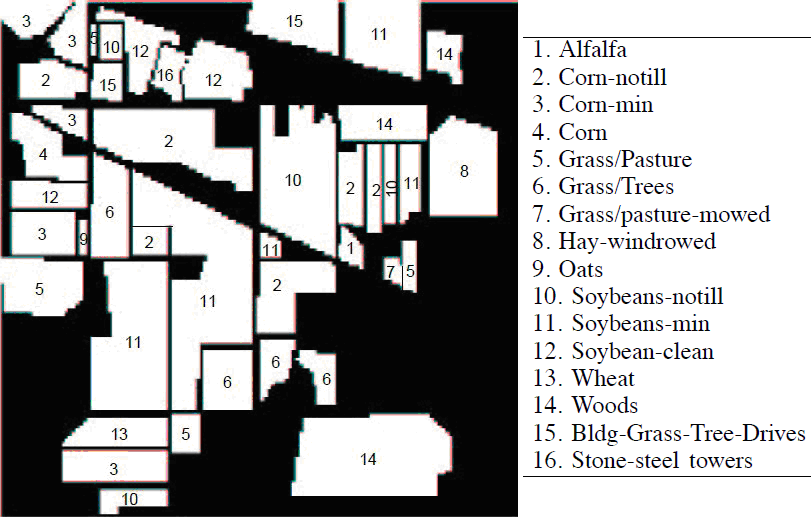}
\caption{The Ground Truth map of AVIRIS 92AV3C and the 16 classes }
\label{fig_sim}
\end{figure}

The hyperspectral image AVIRIS 92AV3C contains numbers between 955 and 9406. Each pixel of the ground truth map has a set of 220 numbers (measures) along the hyperspectral image. This numbers (measures) represent the reflectance of the pixel in each band. So the pixel is shown as vector off 220 components. \\
Figure.3 shows the vector pixel’s notion [7]. So reducing dimensionality means selecting only the dimensions caring a lot of information regarding the classes.

\begin{figure}[!h]
\centering
\includegraphics[width=3.5in]{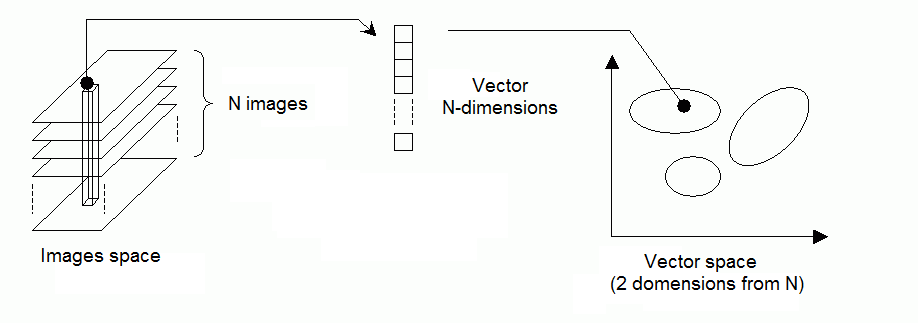}
\caption{The notion of  pixel vector }
\label{fig_sim}
\end{figure}

We can also note that not all classes are carrier of information. In Figure. 4, for
example, we can show the effects of atmospheric affects on bands: 155, 220 and
other bands. This hyperspectral image presents the problematic of dimensionality
reduction.

\section{Mutual Information based feature selection} 
\subsection{Definition of mutual information}

This is a measure of exchanged information between tow ensembles of random variables A and B :
\[
I(A,B)=\sum\;log_2\;p(A,B)\;\frac{p(A,B)}{p(A).p(B)}
\]
Considering the ground truth map, and bands as ensembles of random variables, we calculate their interdependence. \\
Fano [14] has demonstrated that as soon as mutual information of already selected
feature has high value, the error probability of classification is decreasing,
according to the formula bellow:
\[\;\frac{H(C/X)-1}{Log_2(N_c)}\leq\;P_e\leq\frac{H(C/X)}{Log_2}\; \]with :
\[
\;\frac{H(C/X)-1}{Log_2(N_c)}=\frac{H(C)-I(C;X)-1}{Log_2(N_c)}\; \] and :
\[    P_e\leq\frac{H(C)-I(C;X)}{Log_2}=\frac{H(C/X)}{Log_2}\; \]

The expression of conditional entropy\textit{ H(C/X)} is calculated between the ground truth map (i.e. the classes C) and the subset of bands candidates X. Nc is the number of classes. So when the features X have a higher value of mutual information with the ground truth map, (is more near to the ground truth map), the error probability will be lower. But it’s difficult to compute this conjoint mutual information\textit{ I(C;X)}, regarding the high dimensionality [14].

Geo [3] uses also the average of bands 170 to 210, to product an estimated ground truth map, and use it instead of the real truth map. Their curves are similar. This is shown at  Figure 4.

\begin{figure}[!h]
\centering
\includegraphics[width=3.5in]{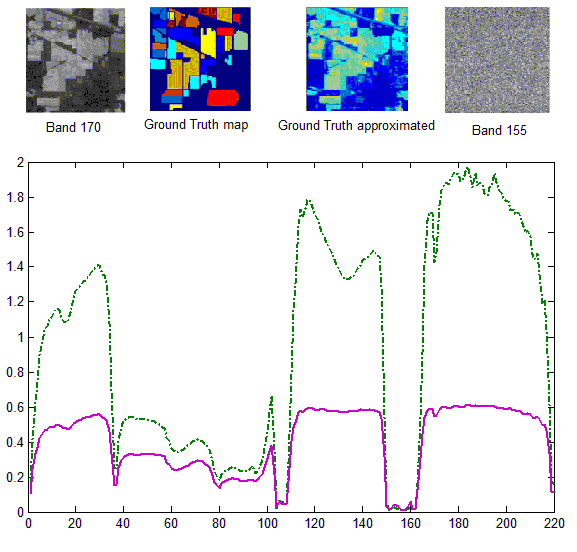}
\caption{Mutual information of AVIRIS  with the Ground Truth map (solid line) and with the ground apporoximated by averaging bands 170 to 210 (dashed line) .}
\label{fig_sim}
\end{figure}

\section{The principe of proposed algorithm}
\subsection{Case of synthetic bands}
\begin{itemize}
	\item{ Band A contains only the class number 11 (Soybeans-min)}
	\item{  Band B contains only the class number 14 (Woods).}
	\item{  Band C contains the bands number 11 and 14.}
\end{itemize}
\begin{figure}[!h]
\centering
\includegraphics[width=3.5in]{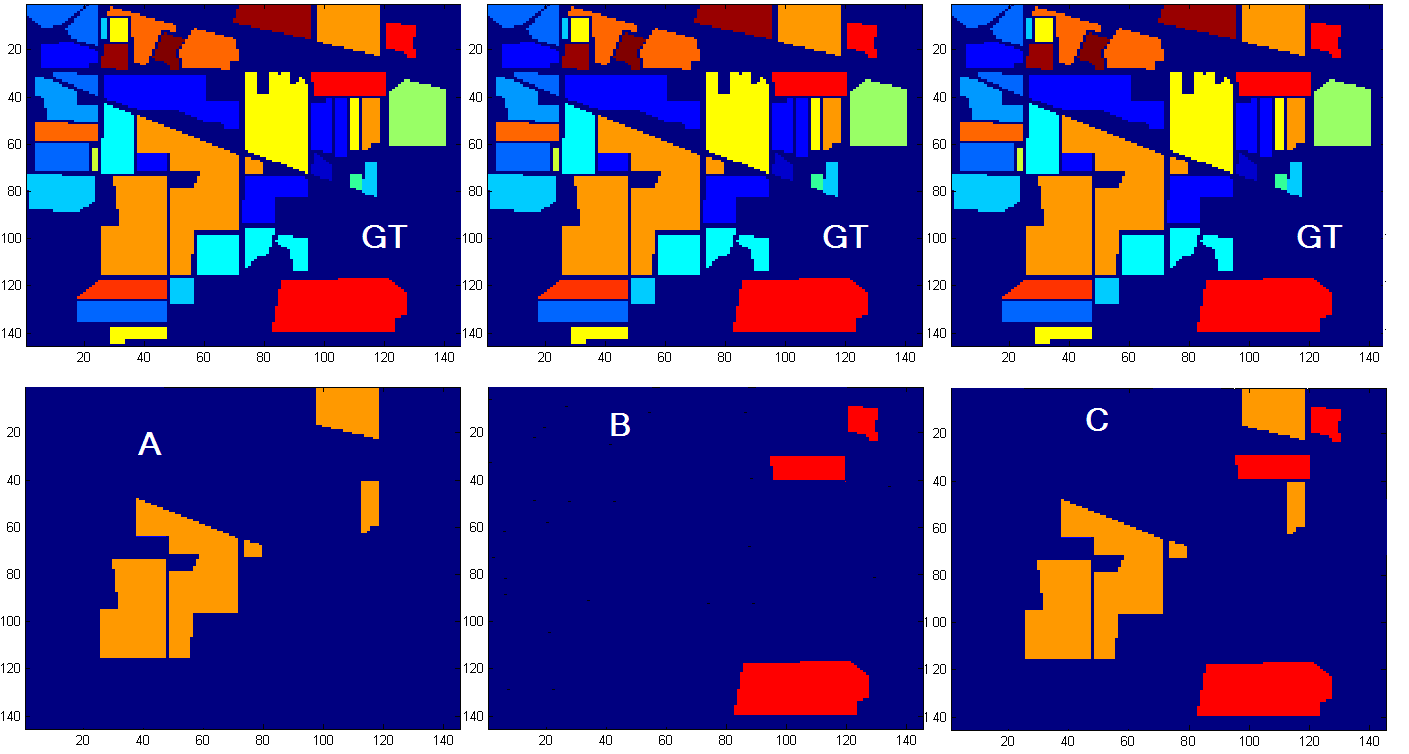}
\caption{Three synthetic bands choised for illustrate the principe of the algorithme.}
\label{fig_sim}
\end{figure}

Now we calculate the mutual information between each of them and the GT. We
compute also the MI between the GT and the superposition of C and B. The
results shown at Table.1

\begin{table}[!h]

\center
 \caption{MI of GT with synthetic bands and the accuracy  of classification in each case }

\begin{tabular}{llllllll}
\noalign{\bigskip}
\hline
\noalign{\smallskip}
${\mathrm Bands }  $& & & & $$  \\
${\mathrm Synthetics }$ & A & B  & C &A,C & B,C & A,B&A,B,C \\
\noalign{\smallskip}
\hline
MI  		      & 0.52 & 0.33 & 0.84 & 0.84 & 0.84& 0.84&0.84 \\
Accuracy(\%)      & 24.6 & 12.6& 36.6 & 36.6 & 36.6 & 36.6&36.6 \\
\hline
\end{tabular}
\end{table}

\subsection{Comments}
Table.I allows us to comment too cases:
\begin{itemize}
	\item{ Case1: The band A and B are superposed to produce band C: The MI of
the estimated reference map C and GT, increases. We have information added.}
	\item{ Case2: The band A and C are superposed to produce an estimated
reference map. We compute its MI with GT. The MI value isn’t change: the band
A added redundant information. It’s the same when we superpose B and C. It’s
also the same when we superpose more redundant bands: A,B and C : no
information added..}
\end{itemize}

\subsection{Partial conclusion}
\textit{{\bf First:} There is an important observation: the superposition of bands A and B to
construct C, can be interpreted as construction of an estimated reference map \textit{\textbf{ bay
averaging the latest one and the band candidate to be selected.}}}\\
{\bf Second: We can emit this roll:} \textit{a band is relevant to classification if it
contributes to produce an estimated reference map, that has the mutual
information with the ground truth map increasing, if else it mast be discarded}

\section{ Proposed algorithm}
Our idea is based on this observation: the band that has higher value of Mutual
Information with the ground truth map can be a good approximation of it. So we
note that the subset of selected bands are the good ones, if thy can generate an
estimated reference map, sensibly equal the ground truth map. We generate the
estimated reference map by averaging the latest one and the band candidate. So
it’s a Filter approach [16] [13].\\
Our process of band selection will be equivalent to following steps: we order the
bands according to value of its mutual information with the ground truth map.
Then we initialize the selected bands ensemble with the band that has highest
value of MI. At a point of process, we build an approximated reference map C\_est
by averaging the latest one and the band candidate, and we compute the
MI(C\_est,GT). The latest band added (at those already selected) must make
MI(C\_est,GT) increasing, if else it will be discarded from the ensemble retained.
Then we introduce a complementary threshold Th to control redundancy. So the
band to be selected must make MI increasing by a step equal to Th. The algorithm
following shows more detail of the process:\\

\begin{algorithm}  
\vspace{0.20cm}                    
\caption{: Let SS be the ensemble of bands already selected and S the band
candidates to be selected.SS is initially empty; R the ensemble of bands condidate, it contains initially all bands (1..220). MI is initialized with a value MI$^*$, X the number of
bands to be selected an Th the threshold controlling redundancy:}  
\center        
\label{algo 1}  
\begin{algorithmic}
\STATE 1) Select the first band to initialize C\_est:
\STATE Select  \textit{band index}$_{s}$ $S$=\textit{argmax}$_{s}$  MI(s);
\STATE $SS\gets \textit{S} $;
\STATE $R\gets \textit{R}\setminus\textit{S}$;
\STATE$C\_est0=Band(S)$;
\STATE2) Selection process:
    \WHILE {$|SS| <  X$} 
        \STATE Select  \textit{band index}$_{s}$ $S$=\textit{argmax}$_{s}$  MI(s) and $R\gets \textit{R}\setminus\textit{S}$;

	\STATE 
\[
C\_est=\frac{C\_est0+Band(S)}{2};	 //C\_est=Build_estimated\_C
\]

        \STATE  $MI= Mutual\_Information(GT,C\_est)$
            \IF {$MI>MI^*+Threshold$}
                \STATE $MI^*=MI$;  
	     \STATE $C\_est0=C\_est$; 
	     \STATE	$SS\gets \textit{SS} \cup \textit{S}$ ; 
	  \ENDIF
\ENDWHILE
\end{algorithmic}
\end{algorithm}
\endcenter

\section{Results and Discussion}

We apply this algorithm on the hyperspectral image AVIRIS 92AV3C [1], 50
the labeled pixels are randomly chosen and used in training; and the other 50
used for testing classification [3]. The classifier used is the SVM [5] [12] [4].\\
We had to choice negatives values of Th. It means that is impossible to increase
accuracy of classification without allowing redundancy.\\

\subsection{Results}
Table.2 shows the results obtained for several thresholds. We can see the
effectiveness selection bands of our algorithm, and the important effect of
avoiding redundancy.\\

\begin{table}[!h]

\center
\caption{Results illustrate elimination of Redundancy using algorithm proposed, for thresholds ($Th$)}
\begin{tabular}{lllllll}
\noalign{\smallskip}\noalign{\smallskip}
\hline
\noalign{\smallskip}
${\mathrm Bands }  $& & & & $Th$  \\
${\mathrm retained }$ & -0.02& -0.01  & -0.005 &-0.004 & -0.0035 & 0 \\
\noalign{\smallskip}

\hline
2   & 47.44& 47.44 & 47.44 & 47.44 & 47.44 & 47.44 \\
3   & 47.87 & 47.87 & 47.87 & 47.87 & 47.87 & 48.92 \\
4   & 49.31 &49.31 &49.31 & 49.31 & 49.31 & - \\
12 & 56.30 & 56.30 & 56.30 & 56.30& 60.76 & - \\
14 & 57.00 & 57.00 & 57.00 & 57.00& 61.80 & - \\
18 & 59.09 & 59.09 & 59.09 & 59.09& 63.00 & - \\
20 & 63.08 & 63.08 & 63.08 & 63.53 & - & -\\
25 & 66.12 & 64.89 & 64.89 & 65.38& - & - \\
30 & 73.54 & 70.72 & 70.72 & 67.68& - & - \\
33 & 73.72 & 74.79 & 75.65 & - & - & - \\
35 & 76.06 & 74.72 & 75.59 & - & - & - \\
36 & 76.49 & 76.60 & 76.19 & - & - & - \\
40 & 78.96 & 79.29 & - & - & - & -\\
45 & 80.85 & 81.01 & - & - & - & - \\
50 & 81.63 & 81.12 & - & - & - & - \\  
53 & 82.27 & 86.03 & - & - & - & - \\
60 & 82.74 & 85.08 & - & - & - & - \\
70 & 86.95 & - & - & - & - & - \\
75 & 86.81 & - & - & - & - & - \\
80 & 87.28 & - & - & - & - & - \\
83 & 88.14 & - & - & - & - & - \\
\hline
\end{tabular}
\end{table}

\subsection{Analysis  and Discussion}

\textbf{Important:}\textit{ When we apply our algorithm on the real data, here AVIRIS 92AV3C,
we note that the can’t increase the accuracy without allowing redundancy bay
negatives thresholds. But the idea is good: we note that the algorithm is selective
and the threshold control effectiveness the redundancy:}\\
\textbf{First:} For the relatively highest threshold values (-0.0035,-0,001,0,+) we obtain a
hard selection: a few more informative bands are selected.\\
\textbf{Second:} For the medium threshold values (-0.01, -0.005, -0.004), some
redundancy is allowed, even if it’s harmful (negative values of thresholds), in
order to made increasing the classification accuracy.\\
\textbf{Tired: }As soon as the threshold value is more negative (-0.02), the redundancy
allowed becomes useless, we have the same accuracy with more bands.
\textbf{Finally:} for the more negative thresholds (for example -4), we allow all bands to
be selected, and we have no action of the algorithm. This corresponds at selection
bay ordering bands on mutual information for numerous thresholds. The
performance is low.\\
We can not here that Hui Wang [15] uses two axioms to characterize feature
selection. Sufficiency axiom: the subset selected feature must be able to reproduce
the training simples without losing information. The necessity axiom "simplest
among different alternatives is preferred for prediction". In the algorithm
proposed, reducing error probability between the truth map and the estimated
minimize the information loosed for the samples training and also the predicate
ones.\\
We not also that we can use the number of features selected like condition to stop
the search. [16].

\textbf{Partial conclusion:}\textit{  The algorithm proposed effectively reduces dimensionality of
hyperspectral images.}









\section{Conclusion} 
In this paper we presented the necessity to reduce the number of bands, in
classification of Hyperspectral images. Then we carried out the effectiveness of
mutual information to select bands able to classify the pixels of ground truth. And
also we insisted on saving the propriety of hyperspectral images regarding the
multispectral images, when we reduce dimensionality. We introduce an algorithm
based on mutual information. To choice a band, it must contribute to reproduce an
estimated ground truth map more closed to the reference map. A complementary
threshold is added to avoid redundancy. So each band retained has to reproduce an
estimated ground truth map more closed to the reference map by a step equal to
threshold even if it caries a redundant information. But the method used her to
estimate the reference map, play an important role: here with averaging bands; we
are constraint to use negative value of threshold; so we allow more redundancy.
We can tell that we conserve the useful redundancy by adjusting the
complementary threshold. This algorithm is a feature selection methodology, and
it’s a Filter approach. It’s less expensive. It can be implemented in real time
applications. This scheme is very interesting to investigate and improve,
considering its performance.


{\bf Received: March, 2012}


\begin{thebibliography}{99} 


\bibitem{IEEEhowto:kopka}
D. Landgrebe, “On information extraction principles for hyperspectral data: A white paper,” Purdue University, West Lafayette, IN, Technical Report, School of Electrical and Computer Engineering, 1997. Téléchargeable ici :
  http://dynamo.ecn.purdue.edu/~landgreb/whitepaper.pdf.

\bibitem{2}
ftp://ftp.ecn.purdue.edu/biehl/MultiSpec/
\bibitem{3}
Baofeng Guo, Steve R. Gunn, R. I. Damper Senior Member, "Band Selection for Hyperspectral Image Classification Using Mutual Information" , IEEE and J. D. B. Nelson. IEEE GEOSCIENCE AND REMOTE SINSING LETTERS, Vol .3, NO .4, OCTOBER 2006.
\bibitem{4}
Baofeng Guo, Steve R. Gunn, R. I. Damper, Senior Member, IEEE, and James D. B. Nelson."Customizing Kernel Functions for SVM-Based Hyperspectral Image Classification",  IEEE TRANSACTIONS ON IMAGE PROCESSING, VOL. 17, NO. 4, APRIL 2008.
\bibitem{5}
Chih-Chung Chang and Chih-Jen Lin, LIBSVM: a library for support vector machines. ACM Transactions on Intelligent Systems and Technology , 2:27:1--27:27, 2011. Software available at http://www.csie.ntu.edu.tw/~cjlin/libsvm.
\bibitem{6}
Nathalie GORRETTA-MONTEIRO , Proposition d'une approche de'segmentation d’images hyperspectrales. PhD thesis. Universite Montpellier II. Février 2009.
\bibitem{7}
David Kernéis." Amélioration de la classification automatique des fonds marins par la fusion multicapteurs acoustiques". Thèse,  ENST BRETAGNE, université de Rennes. Chapitre3, Réduction de dimensionalité et classification,Page.48. Avril 2007.
\bibitem{8}
Kwak, N and Choi, C. "Featutre extraction based on direct calculation of mutual information".IJPRAI VOL. 21, NO. 7, PP. 1213-1231, NOV. 2007 (2007).
\bibitem{9}
 Nojun Kwak and C. Kim,"Dimensionality Reduction Based on ICA Regression Problem". ARTIFICIAL NEURAL NETWORKS-ICANN 2006. Lecture Notes in Computer Science, 2006, Isbn 978-3-540-38625-4,  Volume 1431/2006.
\bibitem{10}
 Huges, G. Information Thaory,"On the mean accuracy of statistical pattern recognizers". IEEE Transactionon Jan 1968, Volume 14, Issue:1, p:55-63, ISSN 0018-9448 DOI: 10.1109/TIT.1968.1054102.

\bibitem{11}
YANG, Yiming, and Jan O. PEDERSEN, 1997.A comparative study of feature selection in text categorization. In: ICML. 97: Proceedings of the Fourteenth International Conference on Machine Learning. San Francisco, CA, USA:Morgan Kaufmann Publishers Inc., pp. 412.420.
\bibitem{12}
Chih-Wei Hsu; Chih-Jen Lin,"A comparison of methods for multiclass support vector machines" ;Dept. of Comput. Sci.  Inf. Eng., Nat Taiwan Univ. Taipei Mar 2002, Volume: 13 I:2;pages: 415 - 425 ISSN: 1045-9227, IAN: 7224559, DOI: 10.1109/72.991427
\bibitem{13}
Bermejo, P.; Gamez, J.A.; Puerta, J.M."Incremental Wrapper-based subset Selection with replacement: An advantageous alternative to sequential forward selection" ;Comput. Syst. Dept., Univ. de Castilla-La Mancha, Albacete; Computational Intelligence and Data Mining, 2009. CIDM '09. IEEE Symposium on March 30 2009-April 2 2009, pages: 367 - 374, ISBN: 978-1-4244-2765-9, IAN: 10647089, DOI: 10.1109/CIDM.2009.4938673
\bibitem{14}
 Lei Yu, Huan Liu,"Efficient Feature Selection via Analysis of Relevance and Redundancy", Department of Computer Science and Engineering; Arizona State University, Tempe, AZ 85287-8809, USA, Journal of Machine Learning Research 5 (2004) 1205-1224.
\bibitem{15}
Hui Wang, David Bell, and Fionn Murtagh, "Feature subset selection based on relevance" , Vistas in Astronomy, Volume 41, Issue 3, 1997, Pages 387-396.
\bibitem{16}
P. Bermejo, J.A. Gámez,  and J.M. Puerta,  "A GRASP algorithm for fast hybrid (filter-wrapper) feature subset selection in high-

\end{thebibliography}
\end{document}